\documentclass[11pt,a4paper]{article}
\usepackage[affil-sl,noblocks]{authblk}
\usepackage[hyperref]{acl2020}
\usepackage{times}
\usepackage{latexsym}
\usepackage{mdframed}
\usepackage{graphicx}
\usepackage{tikzsymbols}

\usepackage{microtype}
\usepackage{amsmath}
\usepackage[absolute,showboxes,overlay]{textpos}
\usepackage{csquotes}
\TPMargin{3pt}

\aclfinalcopy

\title{Automatic Readability Assessment of German Sentences with Transformer Ensembles}

\author[1,$\dag$]{Patrick Gustav Blaneck}
\author[1,2,$\dag$]{Tobias Bornheim}
\author[1,3,$\dag$]{Niklas Grieger}
\author[1,3,*]{Stephan Bialonski}
\affil[1]{Department of Medical Engineering and Technomathematics\authorcr
FH Aachen University of Applied Sciences, Jülich, Germany\authorcr}
\affil[2]{ORDIX AG -- Team Data Science\authorcr}
\affil[3]{Institute for Data-Driven Technologies\authorcr
FH Aachen University of Applied Sciences, Jülich, Germany}
\affil[*]{\textit{bialonski@fh-aachen.de}, $^\dag$Equal contribution}

\date{}

\DeclareMathOperator{\RMSE}{RMSE}

\begin{document}
\maketitle
\begin{abstract}
  Reliable methods for automatic readability assessment have the potential to impact a variety of fields, ranging from machine translation to self-informed learning.
  Recently, large language models for the German language (such as GBERT and GPT-2-Wechsel) have become available, allowing to develop Deep Learning based approaches that promise to further improve automatic readability assessment.
  In this contribution, we studied the ability of ensembles of fine-tuned GBERT and GPT-2-Wechsel models to reliably predict the readability of German sentences.
  We combined these models with linguistic features and investigated the dependence of prediction performance on ensemble size and composition.
  Mixed ensembles of GBERT and GPT-2-Wechsel performed better than ensembles of the same size consisting of only GBERT or GPT-2-Wechsel models.
  Our models were evaluated in the \emph{GermEval 2022 Shared Task on Text Complexity Assessment} on data of German sentences.
  On out-of-sample data, our best ensemble achieved a root mean squared error of $0.435$.
\end{abstract}

\begin{textblock*}{17cm}[-0.1,0](0cm,27.4cm)
  \centering
  \small
  This work was published as part of the conference proceedings of the GermEval 2022\\ Workshop on Text Complexity Assessment of German Text, available online at \href{https://aclanthology.org/volumes/2022.germeval-1/}{https://aclanthology.org/volumes/2022.germeval-1/}. \\
  Please cite as: Patrick Gustav Blaneck, Tobias Bornheim, Niklas Grieger, and Stephan Bialonski. Automatic Readability Assessment of {G}erman Sentences with Transformer Ensembles. In \textit{Proc. GermEval 2022 Workshop on Text Complexity Assessment of German Text: 18th KONVENS 2022}, pages 57--62, Online (2022).
\end{textblock*}

\section{Introduction}

Automatic Readability Assessment (ARA) is a well-known challenge in natural language processing (NLP) research \cite{Martinc2021,Vajjala2021,CollinsThompson2014}.
Systems for reliable readability assessment have the potential to support readers with learning disabilities, inform self-directed learning, or help control the reading level of automatically generated text translations \cite{Vajjala2021}.

The development of methods for text readability assessment may be described in three phases.
(i) Traditional text readability formulas were based on statistical measures of lexical and syntactic features (such as word difficulty and length).
Techniques from NLP further improved upon traditional formulas by incorporating high-level textual features such as semantic and discursive text characteristics \cite{Martinc2021}.
(ii) In the early 21st century, engineered linguistic features were used to train shallow classifiers and regressors from machine learning (such as support vector machines and decision trees)  which further improved prediction accuracy \cite{CollinsThompson2014}.
(iii) The latest phase has been characterized by the advent of large language models (LLMs) developed in the Deep Learning community.
Such neural networks learn features (vector representations of text) automatically from large text corpora during self-supervised pretraining.
Successful network architectures such as BERT \cite{Devlin2018,Rogers2020} or GPT \cite{Radford2018,Radford2019,Brown2020} closely follow the influential transformer model \cite{Vaswani2017} that allows for efficient modeling of long-range correlations in texts.
By combining representations derived from BERT with linguistic features, recent studies observed increased accuracy in assessing the readability of English texts \cite{Lee2021,Imperial2021}.

Training large language models requires large text corpora, a prerequisite that is difficult to meet in languages with fewer resources (compared to English) such as German.
Thus, most approaches to assess the readability of German texts have been based on linguistic features and traditional models from statistical learning such as polynomial regression, support vector machines, or random forests \cite{Hancke2012,Weiss2018,Naderi2019b,Weiss2021}.
Only recently, large language models have become available for German, most notably GBERT \cite{Chan2020}, which is based on BERT, and \mbox{GPT-2-Wechsel} \cite{Minixhofer2021} which was derived from the English GPT-2 model \cite{Radford2019}.
It is largely unknown to which extent these German language models can improve the automatic readability assessment of German texts.

In this contribution, we investigate the ability of ensembles of GBERT and GPT-2-Wechsel models to assess the readability of German sentences.
We combine these models with traditional linguistic features and evaluate our approach on a recently published dataset of German sentences \cite{Naderi2019a}.
Inspired by previous work on ensembling large language models \cite{Risch2020,Bornheim2021}, we studied the dependence of model accuracy on the number of ensemble members and ensemble composition.
Finally, we describe the models that were evaluated in the \emph{GermEval 2022 Shared Task on Text Complexity Assessment} \cite{Mohtaj2022}.
The implementation details of our experiments (Team \enquote{AComplexity}) are available online\footnote{\url{https://github.com/dslaborg/tcc2022}}.

\section{Data and tasks}
\label{sec:data_and_tasks}

The dataset consisted of 1000 labeled sentences \cite{Naderi2019a} and was provided by the organizers of the \emph{GermEval 2022 Shared Task on Text Complexity Assessment} \cite{Mohtaj2022}.
The sentences were drawn from 23 Wikipedia articles. 250 of these sentences were manually simplified by native German speakers \cite{Naderi2019a}.

The scores (labels) were obtained via an online survey system.
Participants were asked to rate the complexity, understandability, and lexical difficulty of the sentences on a 7-point Likert scale. On this scale, 1 denotes the lowest and 7 the highest possible value \cite{Naderi2019a}.
In total, 10650 valid sentence ratings were collected, distributed among the 1000 sentences.

\begin{figure}[t]
    \centering
    \includegraphics[width=\linewidth]{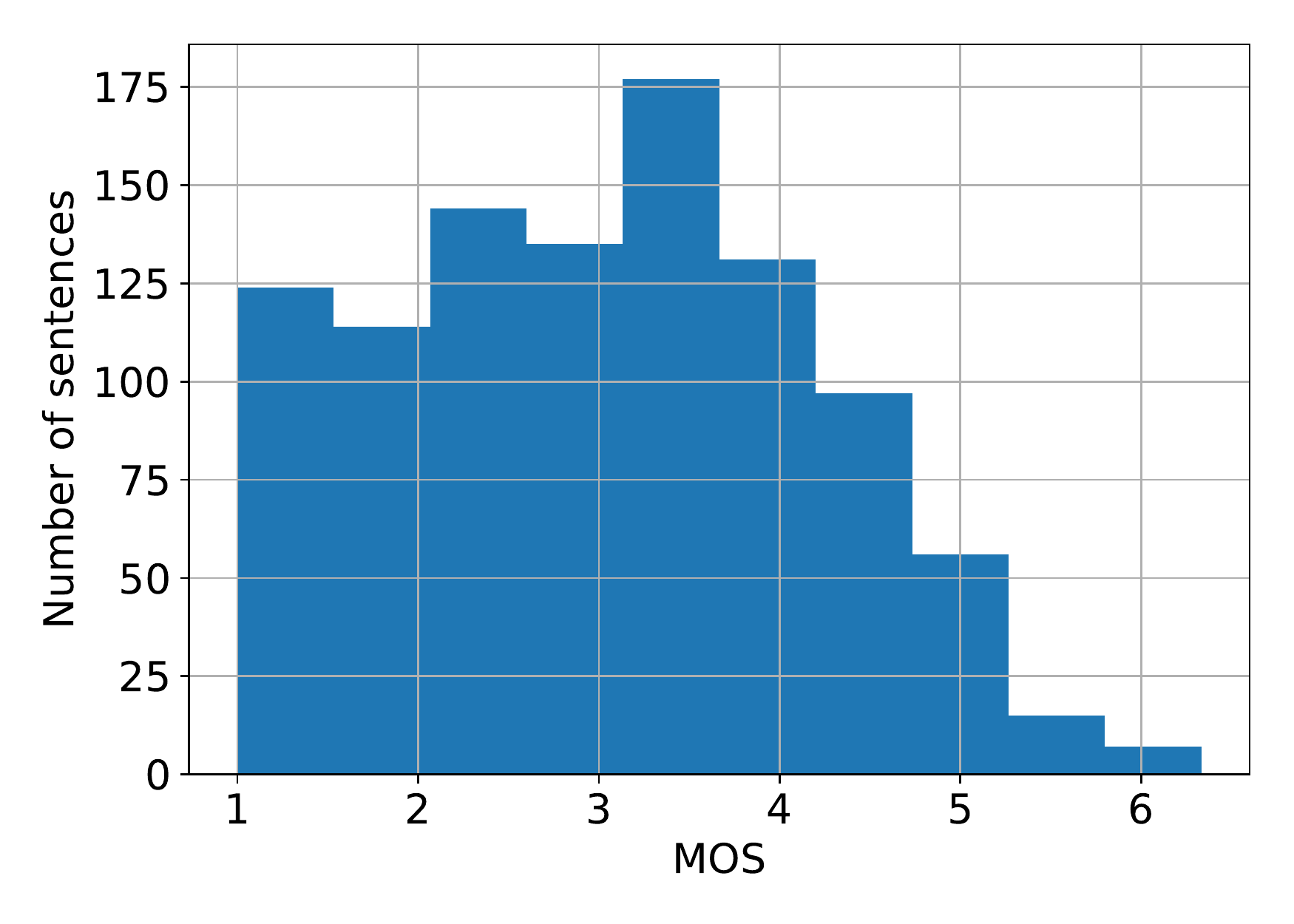}
    \caption{Histogram of Mean Opinion Scores (MOS) for the sentences in the dataset.}
    \label{fig:data_histogram}
\end{figure}

Following a data screening procedure, 5 to 18 ratings per sentence were deemed valid and then used to calculate the arithmetic mean, called the \emph{Mean Opinion Score (MOS)}, of each metric \cite{Naderi2019a}.

\begin{figure}[t]
    \begin{mdframed}{
            \small
            Bei der Tour de France liegt die höchste Durchschnittsgeschwindigkeit eines Fahrers bei 41 km/h. (MOS: 1.5)\\
            \\
            Für die Union resultiert daraus sowohl ein Akzeptanzproblem bei den EU-Bürgern, denen \enquote{Brüssel} immer undurchsichtiger erscheint, als auch die mit dem Mitgliederwachstum verbundene Schwierigkeit, im bestehenden Institutionengefüge die Arbeits- und Handlungsfähigkeit der einzelnen Organe zu gewährleisten. (MOS: 6.33)
        }
    \end{mdframed}
    \caption{Samples (German sentences) from the dataset of the \emph{GermEval 2022 Shared Task on Text Complexity Assessment}. Numbers in parentheses denote text complexity scores.}
    \label{fig:example_data}
\end{figure}

The shared task was to predict the MOS of the text complexity of German sentences.
Since the MOS was defined as a decimal value (see figure~\ref{fig:example_data}), we approached this task as a regression problem.
The distribution of complexity scores (see figure~\ref{fig:data_histogram}) suggests that complex sentences are much less common within the dataset than simpler ones.
Following previous work, we considered text complexity as a proxy of text readability \cite{Wray2013}.

\section{Methods}
\label{sec:methods}

\subsection{Preprocessing and data splits}
\label{ssec:preprocessing}

\emph{Preprocessing.}
All datasets (training, validation, and test data) were preprocessed in the same way.
First, we cleaned up all sentences by removing the leading and trailing quotation marks that were added by the CSV format to mask sentences containing comma separators.
In the next step, all sentences were tokenized with model-specific tokenizers and padded to a uniform length of 128 tokens.

\emph{Data splits.}
During the model exploration phase, models were evaluated with a 5-fold cross validation scheme (each of the five folds contained 20\% of the randomly shuffled training data).
Additionally, we randomly selected 10\% of the data in the training folds (i.e., 8\% of the whole training data) as an \emph{early stopping set} (see section~\ref{sec:training}).
Thus, all models in the model exploration phase were trained on 72\% of the training data.

To optimize model fitting, the final models that were submitted to the \emph{GermEval 2022 Shared Task on Text Complexity Assessment} were retrained on all available training data, aside from a small dataset that was used for \emph{early stopping}.
The \emph{early stopping set} consisted of 7.5\% of the training data and consequently, all final models were trained on 92.5\% of the training data.

\subsection{Readability Features}
\label{ssec:features}

We incorporated various traditional features in the training of our models that are commonly used in text readability and complexity assessment tasks.
The features were generated using two publicly available libraries \cite{VanCranenburgh2019, Proisl2022} and include simple sentence-based measures such as sentence length and punctuation as well as more complex measures such as word rarity.
Furthermore, we included some customized features based on the number of words in a sentence that exceed a given amount of characters.
To increase the amount and variety of the available features, we translated all sentences to English and calculated the features for the original German sentences as well as the English translations.
In total, 154 features were created for each sentence.

\subsection{Models}
\label{ssec:models}
We studied two German language models.
The GBERT model~\cite{Chan2020} is based on the BERT architecture~\cite{Devlin2018}.
We used model weights of the pretrained \emph{gbert-large}\footnote{\url{https://huggingface.co/deepset/gbert-large}} variant, which includes a tokenizer with a vocabulary size of 31000 case-sensitive tokens, has approximately 336 million parameters and a hidden state size of 1024.
Each tokenized sentence was prepended with a classification token that was used for the \mbox{\emph{next sentence prediction}} task during pretraining~\cite{Devlin2018}.

The second model is a German \mbox{GPT-2-Wechsel} model \cite{Minixhofer2021} based on the \mbox{GPT-2} architecture~\cite{Radford2019}.
We used model weights of a pretrained \emph{gpt2-xl-wechsel-german}\footnote{\url{https://huggingface.co/malteos/gpt2-xl-wechsel-german}} variant that was derived from the GPT-2-XL\footnote{\url{https://huggingface.co/gpt2-xl}} model \cite{Radford2019} using the WECHSEL method~\cite{Minixhofer2021}.
The tokenizer has a vocabulary size of 50000 case-sensitive tokens, while the model has roughly 1.5 billion parameters and a hidden state size of 1600.
Since GPT-like models are usually not used for regression tasks, we needed to adjust the tokenizer as follows.
First, we introduced a padding token that was used to pad all sentences to a uniform length of 128 tokens (see section~\ref{ssec:preprocessing}).
Second, we put a \emph{beginning of sequence} token in front and added an \emph{end of sequence} token to the end of every tokenized sentence.

For each transformer model, we employed two different multi-layer perceptron models (MLP) as regression heads.
The first MLP was used to finetune the transformer models on the given training data and did not use the manually created readability features (see section~\ref{ssec:features}).
The second MLP was used after finetuning to incorporate the readability features and consisted of a fully connected layer, followed by ReLu activations and an output layer with one neuron and a linear activation for regression.
The input vector for the second MLP consisted of the output of the last hidden state of the respective transformer model and 154 readability features calculated for each sentence.

\subsection{Training}
\label{sec:training}

\emph{Evaluation score.}
To assess the prediction performance of each model, we calculated the \emph{root mean squared error} (RMSE),

\begin{equation*}
    \mbox{$\RMSE = \sqrt {\frac{1}{N} \sum_{i = 1}^{N} (\hat{y}_i - y_i)^2}$},
\end{equation*}

where $y_i$ denotes the true readability score, $\hat{y}_i$ the predicted readability scores, and $N$ the number of samples in the dataset.
During model exploration, the RMSE was determined for each validation fold of the 5-fold cross validation scheme. We considered the average of these RMSE values as an indicator of model performance.

\emph{Training scheme.} The training was carried out in two phases. In the first phase,
we added a regression head to each model, used an AdamW optimizer~\cite{Loshchilov2019} with a batch size of 16 and a learning rate of \mbox{$\eta=5\cdot 10^{-5}$} with a linear warmup on the first 30\% of the training steps from $0$ to $\eta$.
About every half training epoch (every 23 gradient updates during model exploration or every 28 gradient updates when training the submitted models), the models were evaluated on the \emph{early stopping set}.
If the training lasted for 100 epochs or the RMSE did not decrease for five consecutive evaluations, the training was stopped and the model with the lowest RMSE on the \emph{early stopping set} was returned.
This stopping mechanism was not used during the first 300 gradient updates of the training to prevent underfitting.

In the second phase of the training, the regression heads were discarded and the output of the last hidden state for each sentence of the dataset was extracted as follows.
For GBERT, we used the output of the classification token.
For \mbox{GPT-2-Wechsel}, we extracted the output of the \emph{end of sequence} token.
To create a feature vector for each sentence, we combined the output of the respective transformer model with the readability features calculated for each sentence.
We trained a multi-layer perceptron (MLP) with two layers (see~\ref{ssec:models}) with the RMSprop optimizer, a batch size of 16, and a constant learning rate of $\eta=10^{-3}$.
The MLPs were evaluated on the \emph{early stopping set} after each training epoch.
After 5000 epochs or if the RMSE did not decrease for 100 consecutive epochs, the training was stopped and the model with the lowest RMSE on the \emph{early stopping set} was returned.

During inference, to predict a score for a given sentence, a feature vector was created by combining readability features with the output of the finetuned transformer model. The feature vector served as an input to the trained MLP which calculated the readability score.

\emph{Loss functions.} We used the \emph{mean squared error loss} for training all transformer models and MLPs.

\subsection{Ensembling}
\label{ssec:ensembling}

To counteract the effects of overfitting that often occur when training large models on small datasets, we combined our trained models in ensembles \cite{Risch2020,Bornheim2021}.
Ensemble members differed in the initial model weights of the regression heads and the randomly selected \emph{early stopping set}.
We determined the predictions of an ensemble by averaging the predicted scores of the ensemble members.

\subsection{Postprocessing}
When evaluating our ensembles on the provided test set during the final phase of the competition, we found that some trained models predicted readability scores smaller than 1.0 for a few sentences in the test set.
Since the 7-point Likert scale used by the human annotators to score text readability started at a value of 1.0 (see section~\ref{sec:data_and_tasks}), we deemed all predicted values smaller than 1.0 as invalid and removed them in the ensembling process.
Thus, the predictions of an ensemble were created by averaging only the predicted scores larger than 1.0.
We hypothesize that the scores smaller than 1.0 on the test data were caused by a distribution shift in the generated readability features.

\section{Results}

\begin{figure}[t]
    \includegraphics[width=\linewidth]{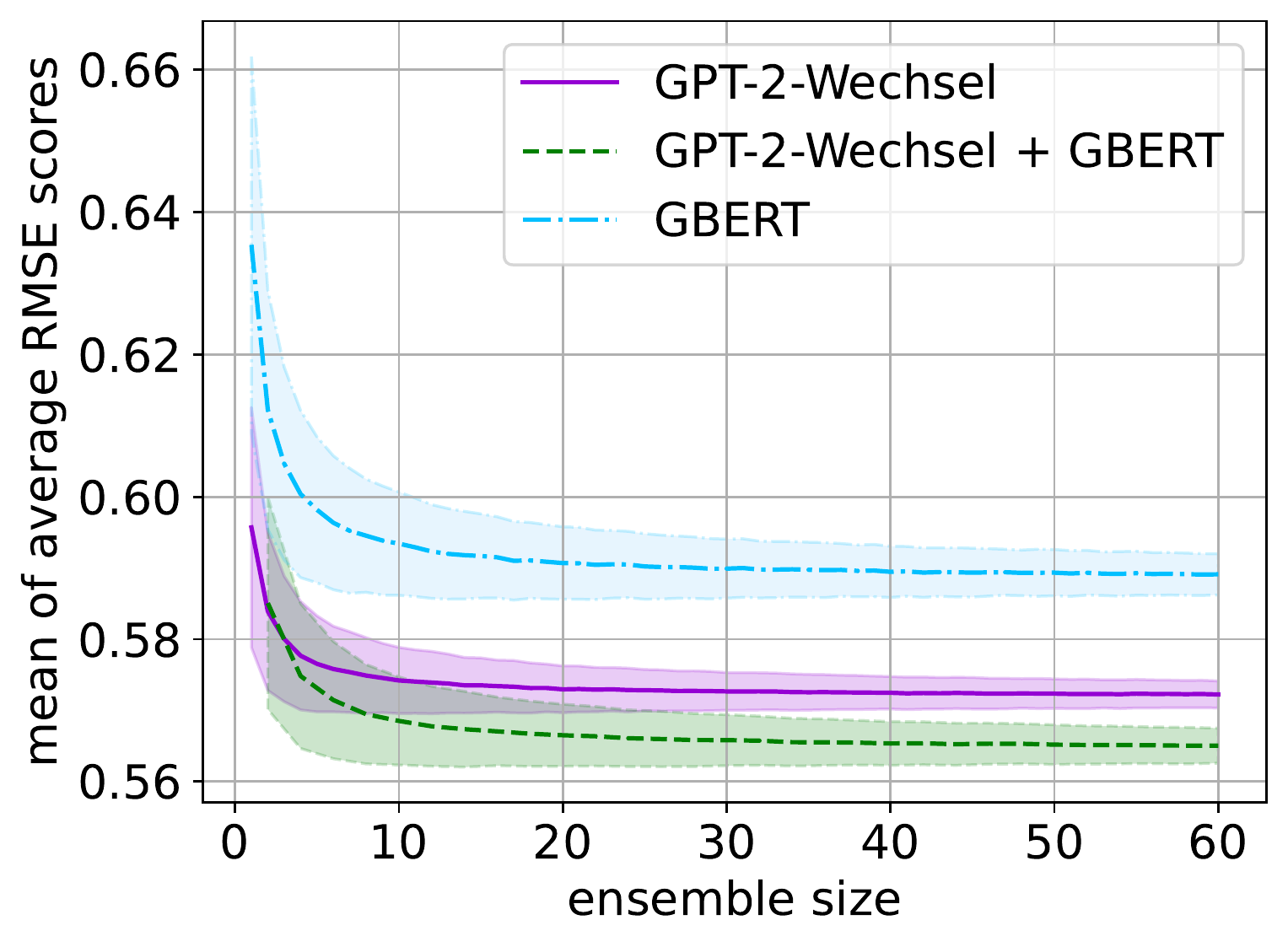}
    \caption{Dependence of the mean of the average root mean squared error (lines) on ensemble size for different ensemble compositions. Standard deviations are shown as shaded areas.}
    \label{fig:ensemble}
\end{figure}

\emph{Model exploration.}
During model exploration we investigated the performance (measured by the average RMSE) of ensembles with different ensemble sizes and compositions.
The ensembles consisted of 1 to 60 models in three different compositions: (i) GBERT models only, (ii) GPT-2-Wechsel models only, (iii) a combination of GBERT and \mbox{GPT-2-Wechsel} models.
In (iii), we combined both model types equally, so that an ensemble of 60 models consisted of 30 GBERT and 30 GPT-2-Wechsel models.

To investigate the dependence of prediction performance on ensemble size, we applied a bootstrapping scheme following \cite{Risch2020,Bornheim2021}.
In total, we trained 100 models each of GBERT and GPT-2-Wechsel on each cross-validation split.
Given a specific ensemble size, we then randomly sampled with replacement 1000 ensembles from the set of trained models and measured the RMSE of each ensemble on each validation fold.
The attained RMSE scores were then averaged over the 5 validation folds, so that we obtained 1000 averaged RMSE scores for each ensemble size.

Figure~\ref{fig:ensemble} shows the mean and standard deviation of the averaged RMSE scores for different ensemble sizes and compositions.
Each ensemble composition benefited from increasing ensemble size, as the mean RMSE decreased considerably up to an ensemble size of 20 models, beyond which the RMSE decreased only slowly.
Increasing the ensemble size also affected the stability of the ensembles' predictions, as can be observed from the decreasing standard deviation of all three ensemble compositions.
Our findings are consistent with previous work \cite{Risch2020,Bornheim2021} which reported improvements in predictive performance when increasing ensemble sizes.

Furthermore, figure~\ref{fig:ensemble} shows large differences in the performance of the three ensemble compositions.
The ensemble that consisted of only GBERT models performed the worst with a mean RMSE of 0.589 at ensemble size 60.
Using \mbox{GPT-2-Wechsel} models instead of GBERT models reduced the mean RMSE to 0.572, and combining both model types in a mixed ensemble of 30 GPT-2-Wechsel and 30 GBERT models further improved the scores to 0.565.

\emph{Submitted models.}
Based on our results in the model exploration phase, we decided to submit two different ensembles in the final phase of the competition: (i) an ensemble of 340 GPT-2-Wechsel models and (ii) an ensemble of 100 GPT-2-Wechsel and 100 GBERT models.
We chose not to submit an ensemble of only GBERT models due to the subpar performance observed during model exploration.
All models were fine-tuned using all available training data, aside from a small dataset (7.5\% of the training data) used for early stopping (see section~\ref{sec:methods}).

On the test data of the shared task, ensembles (i) and (ii) achieved RMSE values of $0.461$ (mapped RMSE: $0.454$\footnote{A linear mapping function was used by the competition organizers; see section 7.3 of the recommendation ITU-T P.1401.}) and $0.442$ (mapped RMSE: $0.435$\footnotemark[5]), respectively \cite{Mohtaj2022}.
Ensemble (ii) ranked 2nd in the competition.

\section{Conclusion}

We studied the ability of ensembles of fine-tuned German language models to reliably predict the readability of German sentences. 
All proposed models also used traditional linguistic features that slightly increased prediction performance (data not shown), consistent with previous reports on text readability assessment of English texts \cite{Imperial2021,Lee2021}. 
We observed mixed ensembles of GBERT and GPT-2-Wechsel to better predict readability scores than ensembles of the same size consisting of only GBERT or GPT-2-Wechsel models. 
Furthermore, prediction accuracy as quantified by the \emph{root mean squared error} decreased with increasing ensemble size, which resembled findings for hate speech classification reported previously \cite{Risch2020,Bornheim2021}.

\appendix


\begin{thebibliography}{25}
  \expandafter\ifx\csname natexlab\endcsname\relax\def\natexlab#1{#1}\fi

  \bibitem[{Bornheim et~al.(2021)Bornheim, Grieger, and Bialonski}]{Bornheim2021}
  Tobias Bornheim, Niklas Grieger, and Stephan Bialonski. 2021.
  \newblock \href {http://arxiv.org/abs/2109.03094} {{FHAC} at {G}erm{E}val 2021:
  {I}dentifying german toxic, engaging, and fact-claiming comments with
  ensemble learning}.
  \newblock \emph{CoRR}, abs/2109.03094.

  \bibitem[{Brown et~al.(2020)Brown, Mann, Ryder, Subbiah, Kaplan, Dhariwal,
        Neelakantan, Shyam, Sastry, Askell, Agarwal, Herbert{-}Voss, Krueger,
        Henighan, Child, Ramesh, Ziegler, Wu, Winter, Hesse, Chen, Sigler, Litwin,
        Gray, Chess, Clark, Berner, McCandlish, Radford, Sutskever, and
        Amodei}]{Brown2020}
  Tom~B. Brown, Benjamin Mann, Nick Ryder, Melanie Subbiah, Jared Kaplan,
  Prafulla Dhariwal, Arvind Neelakantan, Pranav Shyam, Girish Sastry, Amanda
  Askell, Sandhini Agarwal, Ariel Herbert{-}Voss, Gretchen Krueger, Tom
  Henighan, Rewon Child, Aditya Ramesh, Daniel~M. Ziegler, Jeffrey Wu, Clemens
  Winter, Christopher Hesse, Mark Chen, Eric Sigler, Mateusz Litwin, Scott
  Gray, Benjamin Chess, Jack Clark, Christopher Berner, Sam McCandlish, Alec
  Radford, Ilya Sutskever, and Dario Amodei. 2020.
  \newblock \href
  {https://proceedings.neurips.cc/paper/2020/hash/1457c0d6bfcb4967418bfb8ac142f64a-Abstract.html}
  {Language models are few-shot learners}.
  \newblock In \emph{Advances in Neural Information Processing Systems 33: Annu.
    Conf. on Neural Information Processing Systems 2020, NeurIPS 2020}, Virtual.

  \bibitem[{Chan et~al.(2020)Chan, Schweter, and M{\"{o}}ller}]{Chan2020}
  Branden Chan, Stefan Schweter, and Timo M{\"{o}}ller. 2020.
  \newblock \href {https://doi.org/10.18653/v1/2020.coling-main.598} {German's
    next language model}.
  \newblock In \emph{Proc. 28th Int. Conf. on Computational Linguistics, {COLING}
    2020}, pages 6788--6796, Barcelona, Spain (Online). International Committee
  on Computational Linguistics.

  \bibitem[{Collins-Thompson(2014)}]{CollinsThompson2014}
  Kevyn Collins-Thompson. 2014.
  \newblock \href {https://doi.org/https://doi.org/10.1075/itl.165.2.01col}
  {Computational assessment of text readability: {A} survey of current and
  future research}.
  \newblock \emph{Int. J. Appl. Linguistics}, 165(2):97--135.

  \bibitem[{van Cranenburgh(2019)}]{VanCranenburgh2019}
  Andreas van Cranenburgh. 2019.
  \newblock \href {https://github.com/andreasvc/readability} {Readability}.
  \newblock \url{https://github.com/andreasvc/readability/releases/tag/v0.3.1}.

  \bibitem[{Devlin et~al.(2019)Devlin, Chang, Lee, and Toutanova}]{Devlin2018}
  Jacob Devlin, Ming{-}Wei Chang, Kenton Lee, and Kristina Toutanova. 2019.
  \newblock \href {https://doi.org/10.18653/v1/n19-1423} {{BERT:} {P}re-training
    of deep bidirectional transformers for language understanding}.
  \newblock In \emph{Proc. 2019 Conf. North American Chapter of the Association
    for Computational Linguistics: Human Language Technologies, {NAACL-HLT}
    2019}, volume~1, pages 4171--4186, Minneapolis, MN, USA. Association for
  Computational Linguistics.

  \bibitem[{Hancke et~al.(2012)Hancke, Vajjala, and Meurers}]{Hancke2012}
  Julia Hancke, Sowmya Vajjala, and Detmar Meurers. 2012.
  \newblock \href {https://aclanthology.org/C12-1065/} {Readability
    classification for {G}erman using lexical, syntactic, and morphological
    features}.
  \newblock In \emph{{COLING} 2012, 24th Int. Conf. on Computational Linguistics,
    Proc. Conf.: Technical Papers}, pages 1063--1080, Mumbai, India. Indian
  Institute of Technology Bombay.

  \bibitem[{Imperial(2021)}]{Imperial2021}
  Joseph~Marvin Imperial. 2021.
  \newblock \href {https://aclanthology.org/2021.ranlp-1.69} {{BERT} embeddings
    for automatic readability assessment}.
  \newblock In \emph{Proceedings of the International Conference on Recent
    Advances in Natural Language Processing {(RANLP} 2021), Held Online,
    1-3September, 2021}, pages 611--618. {INCOMA} Ltd.

  \bibitem[{Lee et~al.(2021)Lee, Jang, and Lee}]{Lee2021}
  Bruce~W. Lee, Yoo~Sung Jang, and Jason~Hyung{-}Jong Lee. 2021.
  \newblock \href {https://doi.org/10.18653/v1/2021.emnlp-main.834} {Pushing on
  text readability assessment: {A} transformer meets handcrafted linguistic
  features}.
  \newblock In \emph{Proc. 2021 Conf. on Empirical Methods in Natural Language
    Processing, {EMNLP} 2021}, pages 10669--10686, Punta Cana, Dominican Republic
  (Virtual). Association for Computational Linguistics.

  \bibitem[{Loshchilov and Hutter(2019)}]{Loshchilov2019}
  Ilya Loshchilov and Frank Hutter. 2019.
  \newblock \href {https://openreview.net/forum?id=Bkg6RiCqY7} {Decoupled weight
    decay regularization}.
  \newblock In \emph{7th Int. Conf. on Learning Representations, {ICLR} 2019},
  New Orleans, LA, USA.

  \bibitem[{Martinc et~al.(2021)Martinc, Pollak, and
  Robnik-{\v{S}}ikonja}]{Martinc2021}
  Matej Martinc, Senja Pollak, and Marko Robnik-{\v{S}}ikonja. 2021.
  \newblock \href {https://doi.org/10.1162/coli_a_00398} {Supervised and
    unsupervised neural approaches to text readability}.
  \newblock \emph{Comput. Linguist.}, 47(1):141--179.

  \bibitem[{Minixhofer et~al.(2021)Minixhofer, Paischer, and
        Rekabsaz}]{Minixhofer2021}
  Benjamin Minixhofer, Fabian Paischer, and Navid Rekabsaz. 2021.
  \newblock \href {http://arxiv.org/abs/2112.06598} {{WECHSEL:} {E}ffective
    initialization of subword embeddings for cross-lingual transfer of
    monolingual language models}.
  \newblock \emph{CoRR}, abs/2112.06598.

  \bibitem[{Mohtaj et~al.(2022)Mohtaj, Naderi, and Möller}]{Mohtaj2022}
  Salar Mohtaj, Babak Naderi, and Sebastian Möller. 2022.
  \newblock Overview of the {G}erm{E}val 2022 shared task on text complexity
  assessment of {G}erman text.
  \newblock In \emph{Proceedings of the {G}erm{E}val 2022 Shared Task on Text
  Complexity Assessment of {G}erman Text}, Potsdam, Germany. Association for
  Computational Linguistics.

  \bibitem[{Naderi et~al.(2019{\natexlab{a}})Naderi, Mohtaj, Ensikat, and
  M{\"{o}}ller}]{Naderi2019a}
  Babak Naderi, Salar Mohtaj, Kaspar Ensikat, and Sebastian M{\"{o}}ller.
  2019{\natexlab{a}}.
  \newblock \href {http://arxiv.org/abs/1904.07733} {Subjective assessment of
  text complexity: {A} dataset for {G}erman language}.
  \newblock \emph{CoRR}, abs/1904.07733.

  \bibitem[{Naderi et~al.(2019{\natexlab{b}})Naderi, Mohtaj, Karan, and
  M{\"{o}}ller}]{Naderi2019b}
  Babak Naderi, Salar Mohtaj, Karan Karan, and Sebastian M{\"{o}}ller.
  2019{\natexlab{b}}.
  \newblock \href {https://doi.org/10.1109/QoMEX.2019.8743194} {Automated text
  readability assessment for {G}erman language: {A} quality of experience
  approach}.
  \newblock In \emph{11th Int. Conf. on Quality of Multimedia Experience QoMEX
    2019}, pages 1--3, Berlin, Germany. {IEEE}.

  \bibitem[{Proisl(2022)}]{Proisl2022}
  Thomas Proisl. 2022.
  \newblock \href {https://github.com/tsproisl/textcomplexity} {Linguistic and
    stylistic complexity}.
  \newblock
  \url{https://github.com/tsproisl/textcomplexity/releases/tag/v0.11.0}.

  \bibitem[{Radford and Narasimhan(2018)}]{Radford2018}
  Alec Radford and Karthik Narasimhan. 2018.
  \newblock \href
  {https://cdn.openai.com/research-covers/language-unsupervised/language_understanding_paper.pdf}
  {Improving language understanding by generative pre-training}.

  \bibitem[{Radford et~al.(2019)Radford, Wu, Child, Luan, Amodei, and
        Sutskever}]{Radford2019}
  Alec Radford, Jeff Wu, Rewon Child, David Luan, Dario Amodei, and Ilya
  Sutskever. 2019.
  \newblock \href
  {https://cdn.openai.com/better-language-models/language_models_are_unsupervised_multitask_learners.pdf}
  {Language models are unsupervised multitask learners}.

  \bibitem[{Risch and Krestel(2020)}]{Risch2020}
  Julian Risch and Ralf Krestel. 2020.
  \newblock \href {https://www.aclweb.org/anthology/2020.trac-1.9/} {Bagging
      {BERT} models for robust aggression identification}.
  \newblock In \emph{Proc. 2nd Workshop on Trolling, Aggression and
    Cyberbullying, TRAC@LREC 2020}, pages 55--61, Marseille, France. European
  Language Resources Association {(ELRA)}.

  \bibitem[{Rogers et~al.(2020)Rogers, Kovaleva, and Rumshisky}]{Rogers2020}
  Anna Rogers, Olga Kovaleva, and Anna Rumshisky. 2020.
  \newblock \href {https://transacl.org/ojs/index.php/tacl/article/view/2257} {A
  primer in {BERT}ology: {W}hat we know about how {BERT} works}.
  \newblock \emph{Trans. Assoc. Comput. Linguistics}, 8:842--866.

  \bibitem[{Vajjala(2021)}]{Vajjala2021}
  Sowmya Vajjala. 2021.
  \newblock \href {http://arxiv.org/abs/2105.00973} {Trends, limitations and open
    challenges in automatic readability assessment research}.
  \newblock \emph{CoRR}, abs/2105.00973.

  \bibitem[{Vaswani et~al.(2017)Vaswani, Shazeer, Parmar, Uszkoreit, Jones,
        Gomez, Kaiser, and Polosukhin}]{Vaswani2017}
  Ashish Vaswani, Noam Shazeer, Niki Parmar, Jakob Uszkoreit, Llion Jones,
  Aidan~N Gomez, {\L}ukasz Kaiser, and Illia Polosukhin. 2017.
  \newblock \href {http://arxiv.org/abs/1706.03762} {Attention is all you need}.
  \newblock In \emph{Annual Conf. Neural Information Processing Systems 2017},
  pages 5998--6008, Long Beach, CA, USA.

  \bibitem[{Wei{\ss} et~al.(2021)Wei{\ss}, Chen, and Meurers}]{Weiss2021}
  Zarah Wei{\ss}, Xiaobin Chen, and Detmar Meurers. 2021.
  \newblock \href {https://aclanthology.org/2021.nlp4call-1.4.pdf} {Using broad
    linguistic complexity modeling for cross-lingual readability assessment}.
  \newblock In \emph{Proc. 10th Workshop on Natural Language Processing for
    Computer Assisted Language Learning (NLP4CALL 2021)}, Link\"oping Electronic
  Conference Proceedings 177, pages 38--54.

  \bibitem[{Wei{\ss} and Meurers(2018)}]{Weiss2018}
  Zarah Wei{\ss} and Detmar Meurers. 2018.
  \newblock \href {https://aclanthology.org/C18-1026/} {Modeling the readability
    of german targeting adults and children: An empirically broad analysis and
    its cross-corpus validation}.
  \newblock In \emph{Proc. 27th Int. Conf on Computational Linguistics, {COLING}
    2018}, pages 303--317, Santa Fe, New Mexico, USA. Association for
  Computational Linguistics.

  \bibitem[{Wray and Janan(2013)}]{Wray2013}
  David Wray and Dahlia Janan. 2013.
  \newblock Readability revisited? {T}he implications of text complexity.
  \newblock \emph{The Curriculum Journal}, 24:553 -- 562.

\end{thebibliography}
\end{document}